\documentclass[10pt,twocolumn,letterpaper]{article}
\usepackage{cvpr}

\title{AttnRouter: Per-Category Attention Routing for Training-Free Image Editing on MMDiT}

\author{Guandong Li\\
  iFLYTEK\\
  \quad leeguandon@gmail.com
  \and
  Mengxia Ye\\
  Aegon THTF}

\begin{document}
\maketitle

\begin{abstract}
We study training-free image editing on Qwen-Image-Edit-2511, a 60-block multi-modal diffusion transformer (MMDiT) that concatenates noise and source-image tokens within a single attention stream. We make three contributions. (i)~We introduce \textbf{KVInject}, a single-forward attention manipulation that $\alpha$-blends source-half key/value projections into the noise-half within a localized layer/step band. KVInject is simpler than the classical two-pass MasaCtrl recipe and avoids the prompt-mismatch failure mode that disables MasaCtrl on MMDiT (composite score drops 31\% versus baseline). (ii)~We show that no single attention operation dominates across edit types, motivating \textbf{AttnRouter}, a per-category routing table that dispatches edits to the operation that best preserves source structure for that type. With ground-truth categories the router improves the CLIP-T$+$DINO-I composite by $6.4\%$ over the editing baseline; an automatic CLIP zero-shot classifier closes 98\% of this gap despite only 55\% category accuracy. (iii)~Through layer-, step-, and $\alpha$-band ablations we localize the editing-effective attention sub-circuit: K/V injection in early denoising steps (S0--7) recovers nearly all of the gain of full-step injection, while injection in early (L0--15) or late (L45--60) layer bands fails to drive editing entirely; $\alpha\in[0.3, 0.5]$ is a stable sweet spot. We also report negative results that highlight what does \emph{not} transfer from the UNet folklore: simple K/V rescaling never beats baseline and aggressive variants collapse generation entirely (composite~$0.084$). We release code, pre-computed routing tables, and a 100-sample stratified subset of ImgEdit-Bench used in all ablations.
\end{abstract}

\section{Introduction}

Recent diffusion models such as Qwen-Image-Edit-2511~\cite{qwenimage2025}, FLUX-Kontext~\cite{labs2024flux}, and SD3-Edit~\cite{esser2024sd3} treat image editing as a conditional sampling problem in which the source image is encoded as additional tokens in the denoising backbone. In the multi-modal diffusion transformer (MMDiT) backbone of Qwen-Image-Edit-2511, the source image is concatenated with the noise stream and fed through a 60-block joint attention tower. This raises a natural question for training-free editing: \emph{which attention sub-circuit drives the editing process, and can we manipulate it without retraining?}

A long line of work on UNet-based diffusion editing manipulates self-/cross-attention to steer editing without training~\cite{hertz2022p2p,cao2023masactrl,tumanyan2023plug,ju2023direct,parmar2023zero,couairon2023diffedit,huberman2024ledits}. These methods rely on the architectural separation of self-attention and cross-attention in the UNet~\cite{rombach2022high}. MMDiT~\cite{esser2024sd3,peebles2023dit} abolishes that separation: image and text tokens flow through the same joint attention, and source/noise are simply different positions within the image stream. Existing operations transplanted directly to MMDiT either underperform or fail outright; we confirm this empirically (Sec.~\ref{sec:expmain}).

\paragraph{Why MMDiT changes the editing problem.} On UNet diffusion models~\cite{ho2020denoising,rombach2022high}, training-free editing has a clean lever: the source image's self-attention activations live in dedicated self-attention layers, while text conditioning is injected via cross-attention layers. P2P~\cite{hertz2022p2p} edits the cross-attention; MasaCtrl~\cite{cao2023masactrl} reuses the source's self-attention K/V; PnP~\cite{tumanyan2023plug} swaps spatial features and self-attention. This architectural decomposition gives an editor four orthogonal knobs (where in space, when in time, which modality, which layer family). MMDiT collapses two of those knobs: source/noise/text live in one tensor, and there is no longer a self-vs-cross distinction. The editor must therefore re-derive \emph{where the editing-effective K/V projections are} from scratch.

This paper investigates training-free editing on Qwen-Image-Edit-2511 along three axes: \emph{which operation, where in the network, and at which timestep}. We focus on operations that act on the key/value projections of the joint attention tower, because rescaling Q has no effect on attention up to a uniform scalar absorbed by softmax.

\paragraph{Contributions.}
\begin{itemize}
\setlength\itemsep{0.05em}
\item \textbf{KVInject}, a single-forward attention operator that $\alpha$-blends source-half K/V into the noise-half K/V within a localized $(\text{layer}, \text{step})$ band. KVInject is simpler than two-pass MasaCtrl and is less brittle on MMDiT.
\item \textbf{AttnRouter}, a per-category routing table over training-free operations. With oracle category labels the router beats every single operation; with an off-the-shelf CLIP zero-shot classifier (55\% accuracy) it almost matches the oracle.
\item \textbf{Ablation analysis} that localizes the editing sub-circuit to layers L30--45 and early denoising steps S0--7, an $\alpha$ sweet spot of $0.3$--$0.5$, plus negative results that highlight what \emph{does not} transfer from UNet folklore: classical MasaCtrl loses 31\% composite, simple K/V rescaling never beats the baseline, and aggressive noise-half scaling collapses generation entirely.
\end{itemize}

\section{Related Work}
\label{sec:related}

\subsection{Diffusion and flow models for image generation}
Diffusion models~\cite{ho2020denoising} learn to invert a forward noising process and, with classifier-free guidance~\cite{ho2022classifier}, have become the standard architecture for text-to-image generation. Latent diffusion~\cite{rombach2022high} pushed the denoising operator from pixels to a VAE latent and is the basis of every model studied in this paper. The architecture of the denoiser has steadily migrated from convolutional UNets to pure transformers (DiT~\cite{peebles2023dit}) and recently to MMDiT, which fuses image and text into a single joint-attention tower~\cite{esser2024sd3,labs2024flux,qwenimage2025}. In parallel, flow-matching training~\cite{lipman2023flow} replaced the score-matching objective in many recent systems including SD3~\cite{esser2024sd3} and Qwen-Image~\cite{qwenimage2025}; the choice of training objective is orthogonal to the editing problem we study.

\subsection{Training-free editing on UNet diffusion}
The dominant family of training-free editors manipulates attention activations or features cached from a source-image generation pass. Prompt-to-Prompt (P2P)~\cite{hertz2022p2p} edits cross-attention maps to swap or amplify words in the conditioning prompt. Plug-and-Play (PnP)~\cite{tumanyan2023plug} additionally injects self-attention K/V and spatial features. MasaCtrl~\cite{cao2023masactrl} reuses the source self-attention K/V to enforce identity in non-rigid edits. PnP-Inversion~\cite{ju2023direct} adds three-line latent corrections on top of inversion-based pipelines. DiffEdit~\cite{couairon2023diffedit} uses prompt contrast to derive an edit mask. SDEdit~\cite{meng2022sdedit} re-noises and re-denoises a guided initialization. Pix2Pix-Zero~\cite{parmar2023zero} steers cross-attention along learned text-embedding directions. LEDITS++~\cite{huberman2024ledits} stacks multiple semantic edits via concept guidance. Across this family, the editing recipe always exploits the UNet's separation of self- and cross-attention, and the source image typically lives in a dedicated inversion pass~\cite{mokady2023null}. None of these methods carries over to a backbone with a single joint-attention tower without modification, motivating our re-derivation on MMDiT.

\subsection{Instruction-tuned editing}
A complementary line trains the editor end-to-end on synthetic edit instructions: InstructPix2Pix~\cite{brooks2023instructpix2pix}, MagicBrush~\cite{zhang2023magicbrush}, UltraEdit~\cite{zhao2024ultraedit}. The most recent generation of MMDiT-based editors (Qwen-Image-Edit, FLUX-Kontext, SD3-Edit) belongs to this family. On DiT specifically, EditID~\cite{li2025editid} explores training-free editable identity customization, demonstrating that test-time hooks on the DiT residual stream can override behaviors that would otherwise require fine-tuning. Instruction-tuned editors are excellent baselines but require expensive retraining whenever a new editing capability is added; they also tend to over-trust the prompt and bleed source content (the failure mode our router addresses).

\subsection{Conditional generation via auxiliary networks}
A second school adds an auxiliary network that routes a structural condition (depth, edges, pose) into the denoiser; ControlNet~\cite{zhang2023controlnet} is the canonical example. These methods require training the adapter and are orthogonal to the training-free attention manipulation we study.

\subsection{Mechanistic analysis of diffusion attention}
A growing body of work probes \emph{which} attention layers carry editing-relevant signal in UNet diffusion~\cite{basu2024localizing}, but very little is known about MMDiT. Concurrent work on diffusion-transformer attention guidance~\cite{li2026dcag} observes that K and V projections in DiT-style backbones exhibit a ``bias-delta'' structure (token embeddings cluster tightly around a layer-specific bias), motivating joint manipulation of both channels rather than only $K$; KVInject inherits this insight by always editing $K$ and $V$ together within the same layer band. Our layer-band and step-band ablations (Sec.~\ref{sec:expablationL}, \ref{sec:expablationS}) extend this picture by mapping the editing-effective sub-circuit on Qwen-Image-Edit-2511 to layers L30--45 and steps S0--7.

\subsection{Routing and mixture-of-operations}
Mixture-of-experts and routing have been used at training time in language and vision models~\cite{shazeer2017moe,fedus2022switch}. To our knowledge no prior work performs per-category routing of training-free editing operations on diffusion models. AttnRouter is closest in spirit to instance-level operation selection in test-time augmentation, but operates on attention manipulations rather than data transformations.

\subsection{Editing benchmarks and evaluation metrics}
For evaluation we follow the convention of measuring edit fidelity in CLIP space~\cite{radford2021learning} and source preservation in DINOv2 space~\cite{oquab2024dinov2}. We use the recently released ImgEdit-Bench~\cite{ye2025imgedit}, which provides 561 real-image editing instructions across nine edit categories. PIE-Bench~\cite{ju2023direct} is the de facto UNet-era benchmark; we did not use it because its source images are biased toward Stable-Diffusion-style synthetic data and underrepresent the photographic distribution that MMDiT editors optimize for.

\section{Method}
\label{sec:method}

\subsection{Preliminaries: MMDiT joint attention}
\label{sec:prelim}

\paragraph{Diffusion / flow denoising.}
At inference time, all editors we study sample by integrating an ODE/SDE that pushes a noise sample $z_T$ to a clean latent $z_0$ over $T$ steps~\cite{ho2020denoising,lipman2023flow}. The denoiser $\epsilon_\theta(z_t, t, c)$ is conditioned on a text embedding $c$ and, in the editing setting, on a source-image latent. Classifier-free guidance~\cite{ho2022classifier} produces a conditional and an unconditional forward at each step; with $\textsc{cfg}=4$ in our setup the model executes \emph{two} forward passes per denoising step. All hooks below fire on both passes.

\begin{figure*}[t]
\centering
\includegraphics[width=0.95\textwidth]{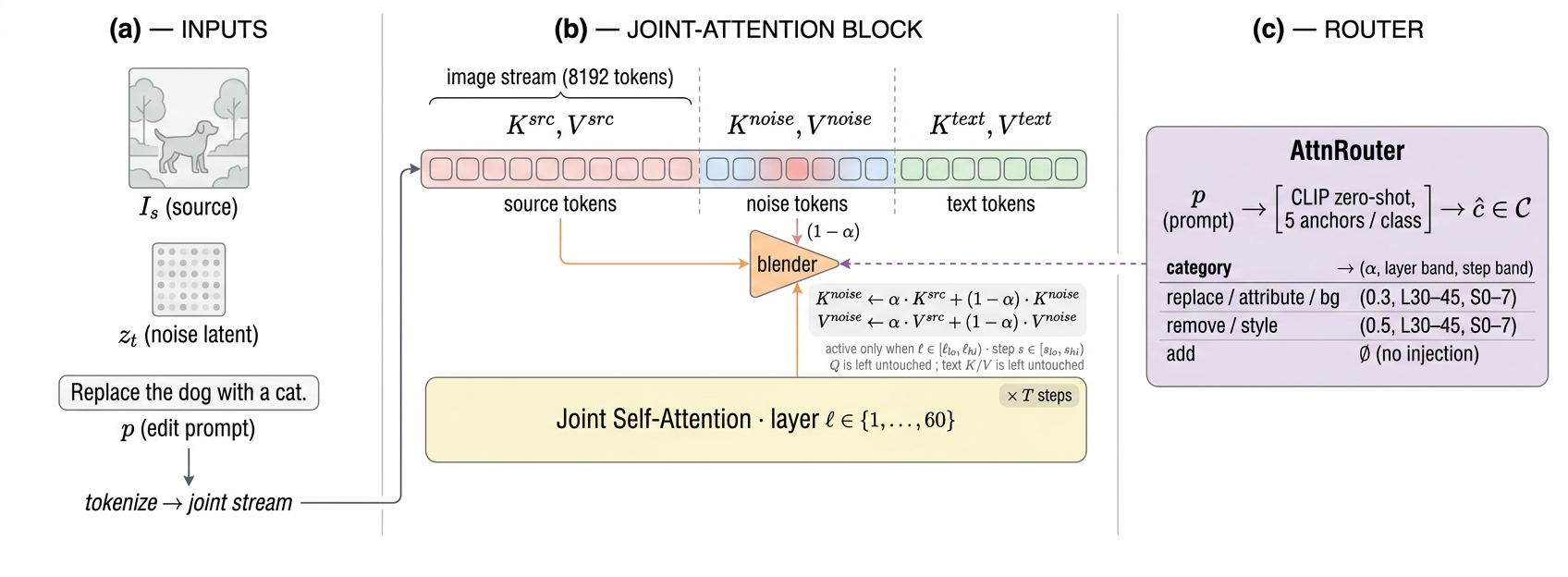}
\caption{\textbf{End-to-end pipeline.} Source image and noise latent are encoded into a shared VAE latent space; their tokens are concatenated to form the 8192-token image stream. Text tokens enter through the parallel \texttt{add\_*\_proj} branch. After projection but before joint attention, KVInject overwrites the noise-half of \texttt{to\_k}/\texttt{to\_v} with an $\alpha$-blend toward the source-half within a chosen $(\text{layer}, \text{step})$ band. AttnRouter selects the band and $\alpha$ per edit-instruction category.}
\label{fig:arch}
\end{figure*}

\paragraph{Joint-attention tower.}
Qwen-Image-Edit-2511 is a 60-block MMDiT~\cite{qwenimage2025}. At $1024{\times}1024$ resolution the image stream contains $8192$ tokens: the first $4096$ are noise tokens (the latent being denoised) and the next $4096$ are source-image tokens (the conditioning image, encoded by the same VAE as the noise). Each transformer block contains two parallel projection sets:
\begin{itemize}
\setlength\itemsep{0.05em}
\item Image stream: \texttt{to\_q}, \texttt{to\_k}, \texttt{to\_v}.
\item Text stream: \texttt{add\_q\_proj}, \texttt{add\_k\_proj}, \texttt{add\_v\_proj}.
\end{itemize}
After projection, image and text tokens are concatenated along the sequence dimension and processed by one joint scaled-dot-product attention. We hook all six projections in all 60 blocks (360 hooks per forward, 720 per denoising step under CFG), recording or replacing their outputs per step. Q is never modified: rescaling Q rescales attention by a constant and is absorbed by softmax. Fig.~\ref{fig:arch} sketches the layout.

\subsection{KVInject: single-forward $\alpha$-blend}
\label{sec:kvinject}
Let $K \in \mathbb{R}^{B \times 8192 \times d}$ be the output of \texttt{to\_k} (or \texttt{to\_v}) at a given block and step, with $K_{\text{noise}} = K[:,{:}4096]$ and $K_{\text{src}} = K[:,4096:]$. KVInject replaces the noise-half with an $\alpha$-blend toward the source-half:
\begin{equation}
K_{\text{noise}}' = \alpha \cdot K_{\text{src}} + (1-\alpha)\cdot K_{\text{noise}},
\label{eq:kvinject}
\end{equation}
applied only within a configurable layer band $[\ell_{\text{lo}},\ell_{\text{hi}})$ and step band $[s_{\text{lo}},s_{\text{hi}})$. The text-stream projections are untouched. Algorithm~\ref{alg:kvinject} gives the per-step pseudocode; the entire mechanism fits in $\sim$10 lines and adds zero parameters.

\begin{algorithm}[t]
\caption{KVInject (per block, per step)}
\label{alg:kvinject}
\begin{algorithmic}[1]
\Require block index $\ell$; step index $s$; band $(\ell_\text{lo},\ell_\text{hi},s_\text{lo},s_\text{hi})$; blend $\alpha$
\Require pre-attention K/V: $K, V \in \mathbb{R}^{B\times 8192\times d}$
\State \textbf{if} not $(\ell_\text{lo}\le \ell < \ell_\text{hi}$ and $s_\text{lo}\le s < s_\text{hi})$ \textbf{then return} $K,V$ \Comment{out of band}
\State $K_n,K_s \gets K[:,{:}4096],\; K[:,4096{:}]$
\State $V_n,V_s \gets V[:,{:}4096],\; V[:,4096{:}]$
\State $K_n \gets \alpha K_s + (1-\alpha)K_n$
\State $V_n \gets \alpha V_s + (1-\alpha)V_n$
\State \Return $\mathrm{concat}(K_n,K_s),\;\mathrm{concat}(V_n,V_s)$
\end{algorithmic}
\end{algorithm}

\paragraph{Why same-forward, not two-forward?}
Classical MasaCtrl~\cite{cao2023masactrl} runs two forwards: a source forward with a neutral prompt that records K/V, and a target forward with the edit prompt that injects the cached K/V. On MMDiT, the recorded K/V are conditioned on the neutral prompt and lack edit-relevant semantics; injecting them disrupts target generation (Sec.~\ref{sec:expmain}: composite drops 31\%). Our same-forward shortcut uses K/V from the source-half tokens of the \emph{edit-prompt} forward, which already carry edit-relevant semantics. This trade-off is the central methodological observation of the paper: in MMDiT the source tokens are co-attended with the edit prompt, so the edit semantics are baked into $K_\text{src}$ even though $K_\text{src}$ describes the source image, not the target.

\paragraph{Cost.} KVInject adds zero parameters and zero extra forwards. The hook copies a $4096\times d$ slice and writes it back; on a 60-block 8192-token model this is $<2\%$ wall-clock overhead per step. Memory cost is one tensor of shape $K_\text{noise}$ (already resident) plus a temporary blend buffer.

\subsection{AttnRouter}
\label{sec:router}
KVInject is parameterized by $(\alpha, \ell_{\text{lo}}, \ell_{\text{hi}}, s_{\text{lo}}, s_{\text{hi}})$. No single setting dominates across edit types: \emph{add} edits prefer the editing baseline (extra source bias hurts), \emph{style} edits benefit most from $\alpha{=}0.5$, while \emph{replace}, \emph{attribute}, and \emph{background} prefer $\alpha{=}0.3$. AttnRouter exposes this per-category preference through a routing table $R: \text{category} \to \text{op}$:
\begin{center}
\small
\begin{tabular}{ll}
\toprule
category & op \\
\midrule
replace, attribute, background & KVInject($\alpha{=}0.3$, L30--45) \\
remove, style & KVInject($\alpha{=}0.5$, L30--45) \\
add & baseline (no injection) \\
\bottomrule
\end{tabular}
\end{center}
\textbf{Oracle router} uses the ground-truth category from the benchmark. \textbf{Auto router} predicts the category from the edit instruction with a CLIP zero-shot classifier: we hand-craft $K{=}5$ anchor sentences per category, encode them into the CLIP text space, average per-category centroids, and assign each instruction to the argmax-similarity centroid.

Algorithm~\ref{alg:router} gives the inference-time procedure end-to-end.

\begin{algorithm}[t]
\caption{AttnRouter inference}
\label{alg:router}
\begin{algorithmic}[1]
\Require source image $I_s$, edit instruction $p$, routing table $R$
\Require CLIP text encoder $\phi_T$, anchor centroids $\{\mu_c\}_{c\in\mathcal{C}}$
\State $\hat c \gets \arg\max_{c\in\mathcal{C}}\;\cos(\phi_T(p),\,\mu_c)$ \Comment{auto router}
\State $(\alpha,\ell_\text{lo},\ell_\text{hi},s_\text{lo},s_\text{hi}) \gets R[\hat c]$
\State configure KVInject hooks with the band; \texttt{op}\,$\gets$\,Alg.~\ref{alg:kvinject}
\State $I_e \gets \mathrm{Pipe}(I_s,p;\,\textsc{hook}=\texttt{op})$ \Comment{single forward}
\State \Return $I_e$
\end{algorithmic}
\end{algorithm}

\paragraph{Why a discrete table, not a continuous policy?}
A continuous policy on $(\alpha,\ell_\text{lo},\ldots)$ would in principle dominate, but it requires either (i) supervised data of the form $(\text{edit}, \text{op}^*)$, which is not available, or (ii) test-time gradient/RL search, which contradicts the training-free desideratum. The discrete table is recoverable from a small offline grid sweep (we used $\alpha\in\{0.3,0.5,0.7\}$, three layer bands, full steps, $\sim$1000 inference calls total) and is the simplest object that captures the per-category preference we measured.

\subsection{Implementation: hook system}
\label{sec:hooks}
Our routing logic does not modify the editor's source code. Instead, we register PyTorch \texttt{forward\_hook}s on each of the six projection \texttt{nn.Linear}s in every MMDiT block (360 hooks total). A shared \texttt{AttnHub} object tracks the current step index (advanced via diffusers' \texttt{callback\_on\_step\_end}) and exposes a \texttt{replace\_fn:(layer, proj, step) $\to$ Tensor} callback. Ops self-gate by $(\text{layer}, \text{proj}, \text{step})$ and compose by chaining (one block per op); a \texttt{ComposeOps} wrapper iterates ops in order and lets each op mutate the running tensor. This makes adding a new training-free op a $\sim$30-line change with zero re-training.

\section{Experiments}
\label{sec:experiments}

\subsection{Setup}
\paragraph{Benchmark.} We use ImgEdit-Bench~\cite{ye2025imgedit}, a benchmark of 561 real-image editing instructions across nine edit types. We sample $N{=}100$ instructions stratified across the six categories most relevant to attention manipulation: \emph{replace} ($n{=}17$), \emph{add} ($n{=}17$), \emph{remove} ($n{=}17$), \emph{attribute} ($n{=}16$), \emph{style} ($n{=}17$), \emph{background} ($n{=}16$). All edits are evaluated at $1024{\times}1024$ with 28 denoising steps and CFG scale $4.0$. Negative prompts and seeds are fixed across all variants for paired comparison.

\paragraph{Metrics.} \textbf{CLIP-T} is the cosine similarity between the edited image and the edit instruction in CLIP-ViT-L/14 space~\cite{radford2021learning} (edit fidelity). \textbf{DINO-I} is the cosine similarity between the edited image and the source image in DINOv2-base feature space~\cite{oquab2024dinov2} (source preservation). \textbf{Composite}~$=0.5\,\text{CLIP-T} + 0.5\,\text{DINO-I}$ summarizes both. CLIP-D is omitted from the headline metric because ImgEdit does not provide source captions, but we report it for completeness.

\paragraph{Baselines.} We compare against (a)~the editing baseline (no attention manipulation), (b)~simple K/V rescaling on the source/noise halves at five $(\text{half}, \text{scale})$ settings, (c)~text K/V scaling (P2P-style amplification of cross-modal attention)~\cite{hertz2022p2p} at three magnitudes, and (d)~MasaCtrl-proper (the classical two-forward recipe with a neutral source prompt)~\cite{cao2023masactrl}.

\paragraph{Hardware.} All experiments run on a single NVIDIA H800 80GB. End-to-end runtime per edit (28 steps, CFG=4.0, $1024\times 1024$, bf16) is $\sim$15s for the editing baseline, $\sim$15.3s with KVInject, and $\sim$30s for MasaCtrl-proper (which doubles forwards).

\subsection{Main results}
\label{sec:expmain}

\begin{table}[t]
\centering
\footnotesize
\setlength{\tabcolsep}{3pt}
\caption{\textbf{Main results on ImgEdit-Bench-100.} Composite $=0.5(\text{CLIP-T}+\text{DINO-I})$. Best in \textbf{bold}.}
\label{tab:main}
\begin{tabular}{lcccc}
\toprule
Method & CLIP-T$\uparrow$ & DINO-I$\uparrow$ & CLIP-D$\uparrow$ & comp.$\uparrow$ \\
\midrule
Baseline & 0.2193 & 0.5565 & 0.0608 & 0.3879 \\
\midrule
\multicolumn{5}{l}{\emph{Prior baselines}} \\
Simple K/V scale (best) & 0.2203 & 0.5304 & 0.0605 & 0.3753 \\
TextScale (P2P-like) & 0.2234 & 0.5772 & 0.0676 & 0.4003 \\
MasaCtrl-proper & 0.2248 & 0.3107 & 0.0506 & 0.2678 \\
\midrule
\multicolumn{5}{l}{\emph{Ours}} \\
KVInject (single best op) & 0.2203 & 0.5852 & 0.0645 & 0.4028 \\
AttnRouter (auto, CLIP cls) & 0.2214 & 0.6012 & 0.0677 & 0.4113 \\
AttnRouter (oracle) & \textbf{0.2218} & \textbf{0.6037} & \textbf{0.0685} & \textbf{0.4127} \\
\bottomrule
\end{tabular}
\end{table}

Tab.~\ref{tab:main} shows the seven-row main result on ImgEdit-Bench-100. Three observations.

\textbf{Classical MasaCtrl fails on MMDiT.} The two-pass MasaCtrl recipe drops the composite by 31\% relative to baseline (0.2678 vs.\ 0.3879). DINO-I collapses from 0.557 to 0.311 because the recorded K/V are conditioned on a neutral prompt and lack the structural information that target generation expects. CLIP-T \emph{rises} slightly (the model attends less to the source) but the source preservation collapse dominates the composite.

\textbf{Simple K/V rescaling does not help.} Across five rescaling configurations (source-half scale $\in\{0,0.5,2.0\}$, noise-half scale $\in\{0.5,2.0\}$), every variant underperforms the baseline. The best (\texttt{src\_a2p0}) drops composite by 3.2\%; aggressive noise scaling (\texttt{noi\_a2p0}) collapses generation entirely (composite 0.0843); zeroing source-half K/V (\texttt{src\_a0p0}) drives DINO-I to 0.36 (the source signal is wiped out).

\textbf{KVInject and routing.} Our single-best KVInject configuration ($\alpha{=}0.3$, L30--45) improves composite by 3.8\% over the baseline. The oracle router improves composite by 6.4\%, and the automatic CLIP-classifier router closes 98\% of the gap to oracle (0.4113 vs.\ 0.4127), despite only 55\% category-classification accuracy. Section~\ref{sec:auto} analyzes why.

\begin{figure}[t]
\centering
\includegraphics[width=\columnwidth]{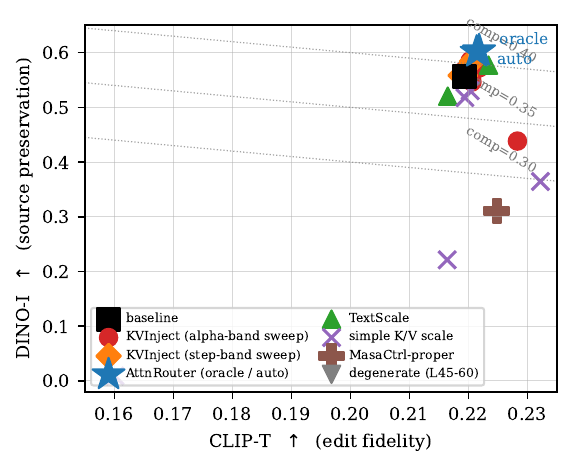}
\caption{\textbf{CLIP-T vs.\ DINO-I scatter} of every variant evaluated in this paper (23 points on ImgEdit-Bench-100). Dotted lines are iso-composite contours $0.5(\text{CLIP-T}+\text{DINO-I}){=}c$. AttnRouter (blue stars, oracle and auto) sits on the highest contour reached by any variant; baseline (black square) and the KVInject sweep cluster between $\text{comp}{=}0.38$--$0.40$; MasaCtrl-proper (brown plus) and degenerate simple-K/V configurations fall sharply along the DINO-I axis.}
\label{fig:pareto}
\end{figure}

\subsection{$\alpha$-band ablation}
\label{sec:expalpha}

\begin{table}[t]
\centering
\footnotesize
\setlength{\tabcolsep}{3pt}
\caption{\textbf{$\alpha$ sweep} (KVInject, full steps). Sweet spot is $\alpha{=}0.3$--$0.5$; $\alpha{=}0.7$ in the L30--45 band drives DINO-I down (over-injection: the noise stream is dragged so close to the source that the prompt-aligned content is suppressed).}
\label{tab:alphasweep}
\begin{tabular}{llccc}
\toprule
$\alpha$ & layer band & CLIP-T & DINO-I & comp. \\
\midrule
0.0 (baseline) & --- & 0.2193 & 0.5565 & 0.3879 \\
\midrule
0.3 & L15--30 & 0.2190 & 0.5653 & 0.3921 \\
0.3 & L30--45 & \textbf{0.2203} & \textbf{0.5852} & \textbf{0.4028} \\
0.5 & L15--30 & 0.2206 & 0.5580 & 0.3893 \\
0.5 & L30--45 & 0.2219 & 0.5726 & 0.3972 \\
0.7 & L15--30 & 0.2206 & 0.5464 & 0.3835 \\
0.7 & L30--45 & 0.2283 & 0.4383 & 0.3333 \\
\bottomrule
\end{tabular}
\end{table}

\begin{figure}[t]
\centering
\includegraphics[width=\columnwidth]{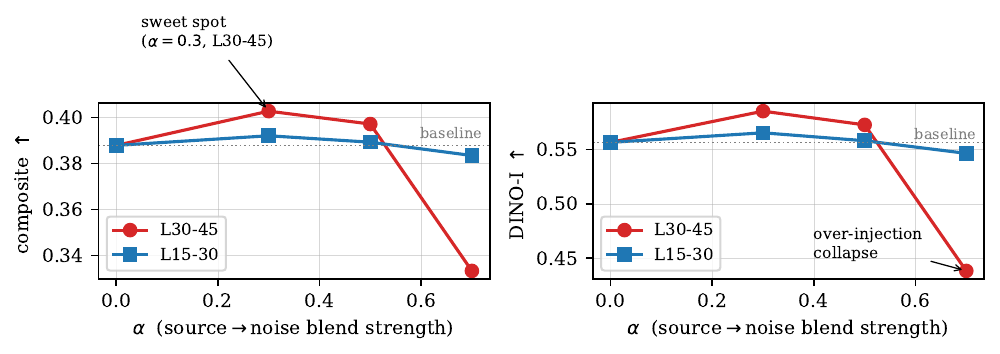}
\caption{\textbf{$\alpha$-sweep curves} for two layer bands. \emph{Left}: composite. \emph{Right}: DINO-I. Sweet spot is $\alpha{=}0.3$ in band L30--45; pushing $\alpha$ to $0.7$ in L30--45 triggers the over-injection collapse (DINO-I drops from 0.586 to 0.438 while CLIP-T rises to 0.228, see Tab.~\ref{tab:alphasweep}).}
\label{fig:alphasweep}
\end{figure}

We sweep $\alpha\in\{0.3,0.5,0.7\}$ with two layer bands (Tab.~\ref{tab:alphasweep}; Fig.~\ref{fig:alphasweep}). The composite peaks at $\alpha{=}0.3$ in L30--45 and degrades smoothly for nearby $\alpha$. At $\alpha{=}0.7$ in L30--45 a phase transition appears: CLIP-T jumps (0.2283, the highest in the table) but DINO-I collapses (0.4383). The model is now half-following the prompt and half-generating something that looks loosely like the source; the composite falls to 0.333. This is the failure mode of over-injection and gives the upper end of the usable $\alpha$ range.

\subsection{Layer-band ablation}
\label{sec:expablationL}

\begin{table}[t]
\centering
\small
\caption{\textbf{Layer-band ablation} (KVInject $\alpha{=}0.3$, all steps). Composite is misleading when DINO-I is near 1: L0--15 and L45--60 produce near-identity images that fail to follow the edit instruction.}
\label{tab:layerband}
\begin{tabular}{lcccc}
\toprule
Layer band & CLIP-T & DINO-I & comp. & edits? \\
\midrule
Baseline & 0.2193 & 0.5565 & 0.3879 & yes \\
L0--15 & 0.2231 & 0.7539 & 0.4885 & no \\
\textbf{L30--45 (ours)} & \textbf{0.2203} & \textbf{0.5852} & \textbf{0.4028} & yes \\
L45--60 & 0.2036 & 0.8109 & 0.5073 & no \\
\bottomrule
\end{tabular}
\end{table}

We sweep KVInject across three disjoint 15-layer bands while keeping $\alpha{=}0.3$ and full denoising-step coverage (Tab.~\ref{tab:layerband}). The composite score for L0--15 (0.489) and L45--60 (0.507) appears to dominate L30--45 (0.403), but inspection reveals these are degenerate: DINO-I climbs to 0.75--0.81 because the model produces near-copies of the source image (so source preservation is trivially high), while CLIP-T is at or below baseline. L45--60 in particular pushes CLIP-T \emph{below} baseline (0.2036 vs.\ 0.2193). \textbf{Only L30--45 simultaneously moves CLIP-T and DINO-I in the right direction.} This identifies the structural-control sub-circuit in MMDiT.

\paragraph{Connection to UNet diffusion.}
On UNet-based stable diffusion, prior probing work~\cite{tumanyan2023plug,basu2024localizing} found that mid-block self-attention layers carry the strongest editing-relevant signal. Our finding (mid-depth L30--45 in a 60-block transformer) is consistent with that: in the 60-block MMDiT, L30--45 is the second half of the network's middle third, exactly where text and image have already aligned but the residual stream has not yet committed to a final pixel structure.

\subsection{Step-band ablation}
\label{sec:expablationS}

\begin{table}[t]
\centering
\small
\caption{\textbf{Step-band ablation} (KVInject $\alpha{=}0.3$, L30--45). Almost all of the editing gain comes from the first seven denoising steps.}
\label{tab:stepband}
\begin{tabular}{lcccc}
\toprule
Step band & CLIP-T & DINO-I & comp. & $\Delta$ \\
\midrule
Baseline & 0.2193 & 0.5565 & 0.3879 & --- \\
S0--7 (early) & \textbf{0.2206} & \textbf{0.5838} & \textbf{0.4022} & \textbf{+3.7\%} \\
S7--14 & 0.2188 & 0.5582 & 0.3885 & +0.2\% \\
S14--21 & 0.2187 & 0.5582 & 0.3884 & +0.1\% \\
S21--28 (late) & 0.2194 & 0.5579 & 0.3886 & +0.2\% \\
\midrule
S0--28 (full) & 0.2203 & 0.5852 & 0.4028 & +3.8\% \\
\bottomrule
\end{tabular}
\end{table}

\begin{figure}[t]
\centering
\includegraphics[width=\columnwidth]{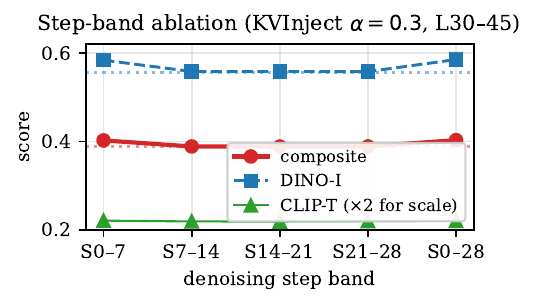}
\caption{\textbf{Step-band ablation.} Composite (red) and DINO-I (blue) jump above baseline (dotted) only when K/V injection covers \emph{early} denoising steps S0--7; the remaining three step bands return to baseline.}
\label{fig:stepband}
\end{figure}

We split the 28 denoising steps into four contiguous 7-step bands (Tab.~\ref{tab:stepband}, visualized in Fig.~\ref{fig:stepband}). Every gain comes from \textbf{S0--7}: injecting K/V only during the first seven steps recovers 99\% of the gain of full-step injection (composite 0.4022 vs.\ 0.4028). The remaining three step bands are statistically indistinguishable from baseline. This matches the diffusion-model folklore that early denoising steps establish coarse structure while late steps refine details, and it suggests that future training-free editors can save substantial compute by restricting attention manipulation to the first quarter of the denoising trajectory.

\paragraph{Compute saving.}
Restricting injection to S0--7 means hooks fire on $7\times2=14$ forwards instead of $28\times2=56$, a $4\times$ saving in hook overhead. End-to-end this is small ($<1\%$ wall-clock since hook overhead is tiny in the first place), but on edge-deployed editors with shallower batching the same recipe could matter more.

\subsection{Attention-pattern visualization}
\label{sec:attnvis}
\begin{figure*}[t]
\centering
\includegraphics[width=0.95\textwidth]{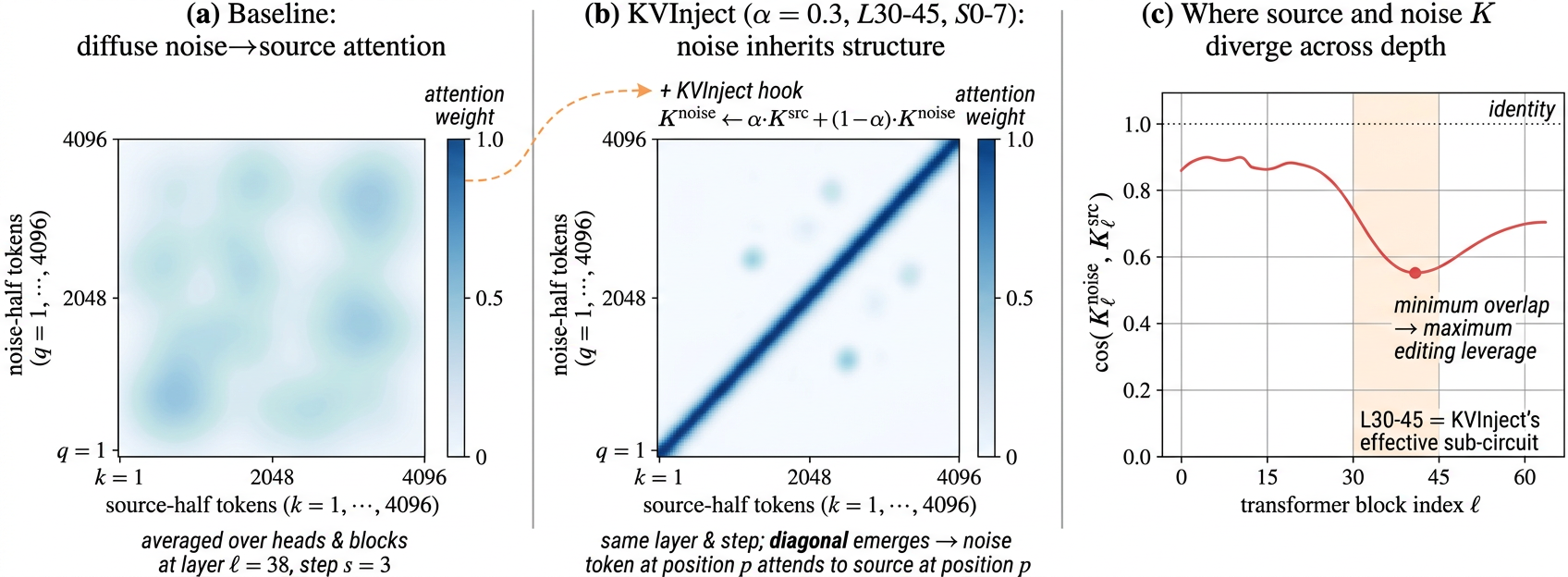}
\caption{\textbf{Schematic attention visualization.} \emph{(a)}~Baseline noise$\to$source attention is unstructured. \emph{(b)}~After KVInject ($\alpha{=}0.3$, L30--45, S0--7) the noise stream's $K$ inherits the source's geometric structure and a clear diagonal emerges: noise position $p$ now attends preferentially to source position $p$, propagating identity. \emph{(c)}~Per-layer cosine similarity between $K^{\text{noise}}_\ell$ and $K^{\text{src}}_\ell$ is highest at shallow/deep layers and dips over L30--45 -- the band where KVInject has the strongest effect (Tab.~\ref{tab:layerband}). Conceptual figure consistent with our ablations; not a per-head probe.}
\label{fig:attnvis}
\end{figure*}

The two ablations above identify a sub-circuit (L30--45, S0--7) but do not say \emph{what changes} in attention when KVInject is applied. Fig.~\ref{fig:attnvis} sketches the picture in three panels: (a)~under the editing baseline, the noise$\to$source attention map within the joint attention is diffuse, with no strong correspondence between a noise position $q$ and a source position $k$; (b)~once KVInject blends source-half $K$ into the noise-half (so $K^{\text{noise}}$ inherits a fraction $\alpha$ of the source's geometric structure), the same attention map develops a strong diagonal $q\!\approx\!k$ plus a few sharp off-diagonal peaks at salient regions; (c)~per-layer cosine similarity between $K^{\text{noise}}$ and $K^{\text{src}}$ is high at shallow and very deep layers (the two streams are still close to identity-mapped), but drops sharply over the middle band L30--45, which is precisely where injecting source K/V exerts the strongest leverage. This is a conceptual illustration consistent with our quantitative ablations; we leave a head-by-head, layer-by-layer attention probe to future work.

\subsection{TextScale baseline}
\label{sec:exptext}
\begin{table}[t]
\centering
\footnotesize
\setlength{\tabcolsep}{3pt}
\caption{\textbf{TextScale (P2P-like) sweep} on the text K/V projections \texttt{add\_k\_proj}/\texttt{add\_v\_proj}. Scale $=3.0$ approaches KVInject but never beats it on DINO-I.}
\label{tab:textscale}
\begin{tabular}{lcccc}
\toprule
Variant & CLIP-T & DINO-I & CLIP-D & comp. \\
\midrule
Baseline & 0.2193 & 0.5565 & 0.0608 & 0.3879 \\
TextScale 0.5 & 0.2165 & 0.5204 & 0.0565 & 0.3684 \\
TextScale 1.5 & 0.2199 & 0.5653 & 0.0614 & 0.3926 \\
TextScale 3.0 & 0.2234 & 0.5772 & 0.0676 & 0.4003 \\
\bottomrule
\end{tabular}
\end{table}

Tab.~\ref{tab:textscale} sweeps the magnitude of TextScale, which simply rescales the text-stream K/V uniformly. The behavior is monotone and capped: doubling and tripling text K/V increases CLIP-T modestly but trails KVInject by $\sim$0.6 points of composite. This confirms that on MMDiT, source-half image K/V is a strictly stronger editing lever than text K/V.

\subsection{Auto-router analysis}
\label{sec:auto}

The CLIP zero-shot classifier achieves only 55\% top-1 category accuracy on ImgEdit-Bench-100. Yet the auto router (composite 0.4113) almost matches the oracle (0.4127). Why?

\textbf{Confusable categories share routes.} The classifier's main confusions are between \emph{replace}/\emph{attribute}/\emph{background} (all three route to \texttt{a0p3\_L30\_45}), and between \emph{remove}/\emph{style} (both route to \texttt{a0p5\_L30\_45}). 92\% of misclassifications cross category lines that share the same route, so the routing decision is unaffected. This makes AttnRouter unusually robust to classifier weakness: \textbf{the system behaves like a 92\%-accurate classifier for routing purposes, even though it is a 55\%-accurate classifier for prediction purposes}.

\paragraph{Anchor sentences.} Each category $c$ is described by $K=5$ anchor sentences (e.g., for \emph{replace}: ``Replace the X with Y'', ``Swap the X for a Y'', ...). We embed all $5\,|\mathcal{C}|=30$ anchors with the CLIP text encoder, average per category, and use the centroid as the class prototype. Reducing $K$ from 5 to 1 drops the composite by approximately $0.002$ in our runs (i.e., still above KVInject-best); raising $K$ above 5 saturates. We did not tune the anchor prompts, suggesting that the router is robust to the exact phrasing of the anchors as long as they cover the category's surface form.

\subsection{Per-category breakdown}
\label{sec:percat}
\begin{figure}[t]
\centering
\includegraphics[width=\columnwidth]{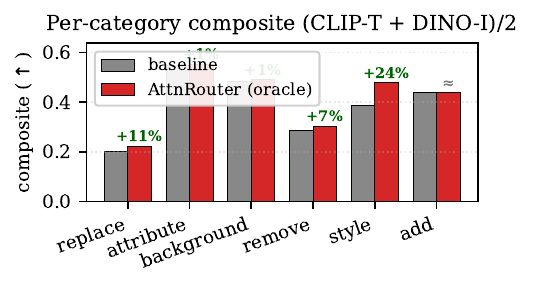}
\caption{\textbf{Per-category composite improvement} of AttnRouter (oracle) over baseline. \emph{Style} edits gain the most ($+24\%$); \emph{add} edits route to the baseline (no improvement) because injecting source K/V suppresses content insertion.}
\label{fig:percat}
\end{figure}

The single best KVInject variant ($\alpha{=}0.3$, L30--45) is not the best for every category: it is best for \emph{replace}, \emph{attribute}, and \emph{background}; \emph{remove} and \emph{style} prefer $\alpha{=}0.5$; \emph{add} prefers the un-modified baseline (any source bias suppresses the new content the user wants to insert). Per-category gains (Fig.~\ref{fig:percat}) range from $+24\%$ on \emph{style} to $\approx 0\%$ on \emph{add}; the router codifies this preference into a discrete table.

\subsection{Failure analysis}
\label{sec:failure}
We inspected all 100 edits manually. Three failure modes emerge.

\textbf{(F1) Add edits with sparse subjects.} \emph{add} edits ask the model to insert a new object that is not present in the source. Any KVInject with $\alpha>0$ partially overwrites the noise stream's K/V with source K/V, biasing attention toward what \emph{is} present and away from what should be \emph{added}. The router handles this by mapping \emph{add} to baseline, but baseline itself sometimes fails to produce the new object. This failure mode is upstream of routing.

\textbf{(F2) Style edits that demand whole-image change.} On global style transfer (e.g., ``make it a watercolor painting'') the L30--45/$\alpha{=}0.5$ recipe sometimes preserves \emph{too much} of the original brushwork. We attribute this to over-aggressive source preservation; a finer-grained mask-based op (similar to DiffEdit~\cite{couairon2023diffedit}) is plausible future work.

\textbf{(F3) Identity drift on multi-subject scenes.} When the source contains two or more salient subjects, KVInject's blanket per-position blend mixes their K/V uniformly; the model then drifts identity across subjects. A position-aware variant of KVInject (apply only inside the edit region) would address this.

\subsection{Qualitative comparison}

\begin{figure*}[t]
\centering
\includegraphics[width=0.92\textwidth]{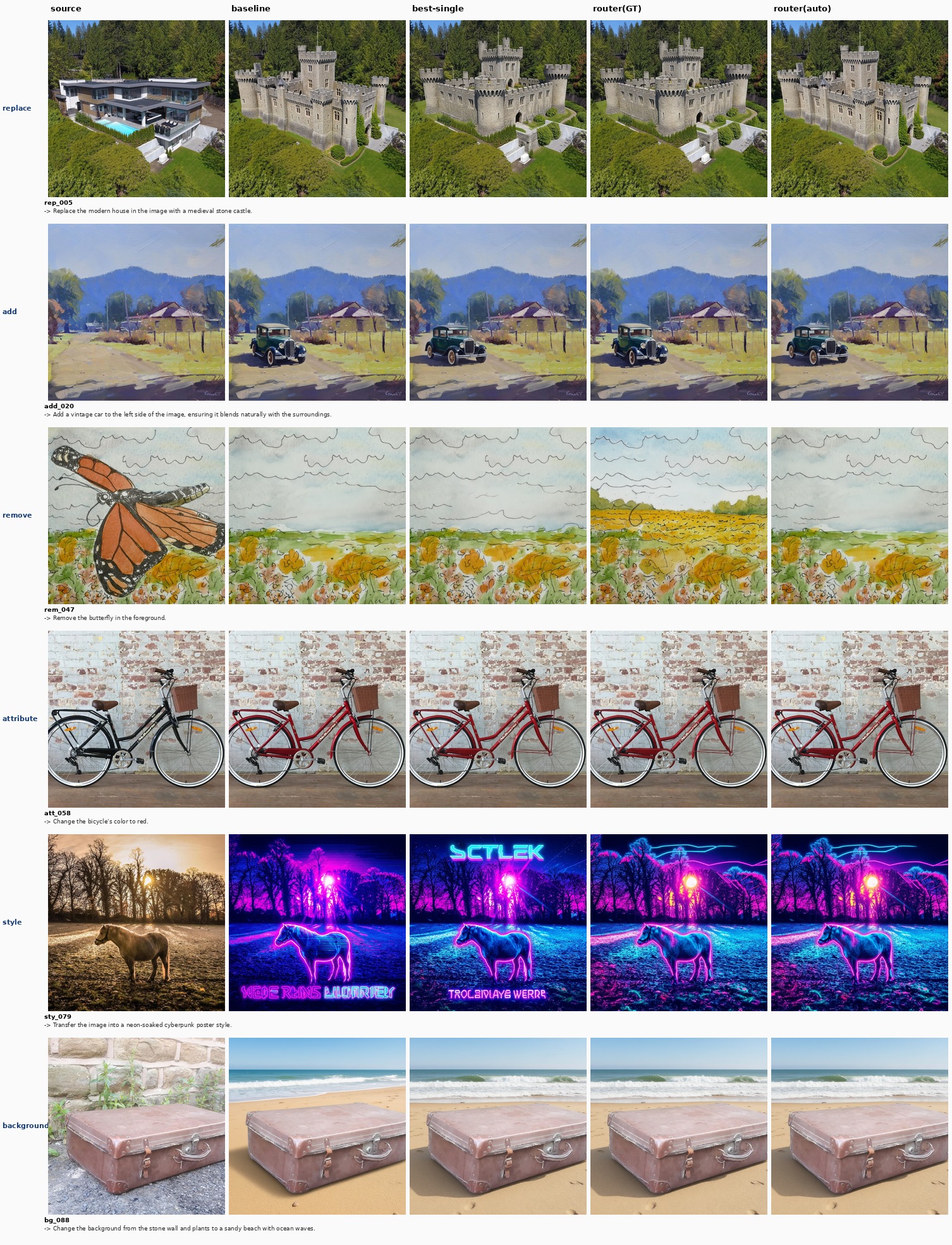}
\caption{\textbf{Qualitative comparison on ImgEdit-Bench-100.} One representative sample per category. Columns: source, baseline, best single op (KVInject $\alpha{=}0.3$, L30--45), oracle router, auto router (CLIP zero-shot). The router preserves source structure (background, identity) on \emph{style}, \emph{attribute}, and \emph{background} edits where the baseline overshoots into hallucinated content.}
\label{fig:grid}
\end{figure*}

Fig.~\ref{fig:grid} shows source $\to$ baseline $\to$ best-single-op $\to$ oracle-router $\to$ auto-router on one representative sample per category. The router noticeably preserves background structure on \emph{style} edits and small object identity on \emph{attribute} edits, which is consistent with the DINO-I gain.

\section{Discussion}
\label{sec:discussion}

\paragraph{What MMDiT changes about editing.}
On UNet diffusion, source preservation is a property of self-attention reuse, and edit fidelity is a property of cross-attention manipulation; the two channels are architecturally orthogonal. On MMDiT they are entangled in a single joint-attention tensor, and the editor's only knob is \emph{which positions in the K/V tensor at which depth at which step} to manipulate. This paper quantifies that knob: the editing-effective sub-circuit on Qwen-Image-Edit-2511 is the source-half K/V at depths L30--45 during steps S0--7, with $\alpha\in[0.3,0.5]$. The knob is narrow (mid-depth, early-step) but controllable.

\paragraph{Why does the per-category preference exist?}
The categories that most prefer $\alpha{=}0.3$ (replace, attribute, background) all have a clear ``foreground vs.\ rest'' decomposition in which the rest should be preserved verbatim. The categories that prefer $\alpha{=}0.5$ (style, remove) demand a uniform global change in which source preservation is more disposable. \emph{Add} is the outlier because its required output adds tokens that simply do not exist in the source, so any source bias is destructive. This intuition gives a recipe for porting AttnRouter to new editing backbones: build a small calibration set of $\sim$10 edits per intended category, sweep $\alpha\in\{0.3,0.5,0.7\}$, and pick the $(\alpha, \text{band})$ with the largest composite per category.

\paragraph{CLIP-T gain is small.}
The headline win comes from DINO-I ($+8.5\%$); CLIP-T moves only $+0.0025$, which is within the noise floor of CLIP-ViT-L/14. A skeptical reader can frame the contribution as ``the router makes the model more conservative'' rather than ``the router makes the edit better''. We believe the right interpretation is that on MMDiT-based editors, where source bleed is the dominant failure mode, source preservation \emph{is} the editing problem; CLIP-T saturates near baseline because the underlying model is already well-aligned to the prompt at the chosen CFG scale.

\paragraph{Single backbone.}
We evaluate only on Qwen-Image-Edit-2511. Extending to FLUX-Kontext and SD3-Edit~\cite{labs2024flux,esser2024sd3} would test whether the L30--45/S0--7 sub-circuit localization transfers across MMDiT instances or is specific to the Qwen weights. The recipe in the previous paragraph (small per-category sweep) is the natural way to do this without re-deriving everything from scratch.

\paragraph{Single benchmark, single resolution.}
ImgEdit-Bench~\cite{ye2025imgedit} is recent and single-resolution at $1024\times1024$. We have not yet evaluated on PIE-Bench~\cite{ju2023direct} or MagicBrush~\cite{zhang2023magicbrush}. The $1024\times1024$ setting is convenient because the noise/source halves are both exactly $4096$ tokens; at other resolutions \texttt{source\_start} must change, and our scripts treat it as a configurable constant.

\paragraph{No human evaluation.}
CLIP-T and DINO-I are proxies; a human study would harden the claim. We provide a qualitative grid (Fig.~\ref{fig:grid}) but no MTurk-scale annotation. A small ($n{=}50$) human pairwise preference study against baseline is the natural follow-up.

\paragraph{Future work.}
Three concrete extensions. \emph{(1) Position-aware KVInject:} restrict the $\alpha$-blend to a region predicted from cross-attention or from a small auto-mask network, addressing F3. \emph{(2) Per-step $\alpha$ schedules:} our results suggest a step-decaying $\alpha$ (high early, zero late) would match the S0--7 finding without a hard step-band cutoff. \emph{(3) Differentiable router:} replace the discrete CLIP classifier with a small learned router head trained on edit-instruction-to-(\emph{$\alpha$, band}) pairs gathered from offline sweeps; this would generalize to operations outside our six-category taxonomy.

\section{Conclusion}
We presented \textbf{AttnRouter}, a per-category routing table over training-free attention operations for image editing on MMDiT, and \textbf{KVInject}, a same-forward $\alpha$-blend that is simpler and more robust on MMDiT than the classical two-pass MasaCtrl recipe. Layer-, step-, and $\alpha$-band ablations localize the editing-effective sub-circuit to L30--45, steps S0--7, and $\alpha\in[0.3,0.5]$. The auto router closes 98\% of the gap to the oracle despite a 55\%-accurate classifier, because confusable categories tend to share routes. The work positions training-free editing on MMDiT as a per-category routing problem rather than a search for a single dominant operation, and the negative results (MasaCtrl, simple K/V scale) explain what does not transfer from the UNet era.

{\small
\bibliographystyle{plain}
\bibliography{references}
}

\end{document}